\documentclass[letterpaper, 10 pt, conference]{ieeeconf}  

\IEEEoverridecommandlockouts                              

\usepackage{cite}
\usepackage{amsmath,amssymb,amsfonts}
\usepackage{algorithmic}
\usepackage{graphicx}
\usepackage{textcomp}
\usepackage{xcolor}
\usepackage{hyperref}

\newcommand{\dynpatch}{\hat{p}_{dyn}}
\newcommand{\dynpatchi}{\hat{p}_{dyn,i}}
\newcommand{\segpatch}{\hat{p}_{seg}}
\newcommand{\segfeat}{\hat{f}_{seg}}
\newcommand{\dynfeat}{\hat{f}_{dyn}}
\newcommand{\binarymask}{\hat{u}_{seg}}

\newcommand{\dynbox}{\hat{x}_{dyn}}
\newcommand{\segbox}{\hat{x}_{seg}}

\newcommand{\predim}{\hat{I}}
\newcommand{\unim}{\hat{I}_{un}}
\newcommand{\synthim}{\hat{I}_{synth}}
\newcommand{\objim}{\hat{I}_{obj}}
\newcommand{\backim}{\hat{I}_{back}}
\newcommand{\im}{I}

\newcommand{\dynmodel}{\mathcal{D}}

\newcommand{\generator}{\mathcal{G}}
\newcommand{\feattopix}{\mathcal{F}}

\newcommand{\segmask}{\hat{m}_{seg}}

\newcommand{\dynmask}{\hat{m}_{dyn}}
\newcommand{\dynmaski}{\hat{m}_{dyn,i}}

\title{\LARGE \bf
Object-centric Video Prediction without Annotation
}

 \author{Karl Schmeckpeper,*$^{1}$ Georgios Georgakis,*$^{1}$ and Kostas Daniilidis$^{1}$
 \thanks{* Denotes equal contribution.}
 \thanks{$^{1}$The authors are with the GRASP Laboratory, Computer and Information Science Department, Univeristy of Pennsylvania, Philadelphia, PA 19104.
         E-Mail: {\tt\small {karls}@seas.upenn.edu}}%
 }

\begin{document}

\maketitle
\thispagestyle{empty}
\pagestyle{empty}

\begin{abstract}
In order to interact with the world, agents must be able to predict the results of the world's dynamics. A natural approach to learn about these dynamics is through video prediction, as cameras are ubiquitous and powerful sensors. Direct pixel-to-pixel video prediction is difficult, does not take advantage of known priors, and does not provide an easy interface to utilize the learned dynamics. Object-centric video prediction offers a solution to these problems by taking advantage of the simple prior that the world is made of objects and by providing a more natural interface for control. However, existing object-centric video prediction pipelines require dense object annotations in training video sequences. In this work, we present Object-centric Prediction without Annotation (OPA), an object-centric video prediction method that takes advantage of priors from powerful computer vision models. We validate our method on a dataset comprised of video sequences of stacked objects falling, and demonstrate how to adapt a perception model in an environment through end-to-end video prediction training.

\end{abstract}


\section{Introduction}

Modeling physical interaction is a fundamental agent skill for interacting with the world.
This is a challenging skill to learn as it requires understanding the scene's dynamics. For object manipulation scenarios, the challenge is exacerbated by the need to understand the environment at the object level, including agent-object and object-object physical interactions.
Addressing this problem using visual sensors offers many advantages.
First, high quality cameras are easily accessible and have low size, weight, and power requirements, allowing them to be included on most robotic platforms.
Second, there is an abundance of existing data available, allowing powerful deep learning models to be trained.
Third, visual observations offer rich information about the environment, including pose, texture, and semantics, that cannot easily be matched by other sensors.

Existing methods typically address this problem via learning an action-conditioned predictive model that infers the changes to the visual scene. These models have been demonstrated mostly through end-to-end deep networks that learn to map pixels and control inputs to future pixels~\cite{finn2017deep}.
This paradigm assumes that the model can implicitly learn to visually segment the objects and infer their motion in the scene in spite of the high dimensionality of pixels from the raw image inputs. 
It does not take advantage of perceptual priors that can be extracted from an observation, forcing the model to function without any aid from existing computer vision methods.

There have been efforts to learn pairwise object interactions by treating a visual scene as a collection of objects, which led to the development of object-centric predictive models. These methods
typically assume that labeled object information is readily available at each future time step either in the form of object locations~\cite{ye2019compositional}, or forces applied on the objects~\cite{fragkiadaki2015learning}.  
However, many robotic agents are required to operate in unstructured real-world environments that exhibit increased visual variability with no access to dense labels and often have to generalize to previously unseen objects.

\begin{figure}[t]
\begin{center}
\includegraphics[width=1\linewidth]{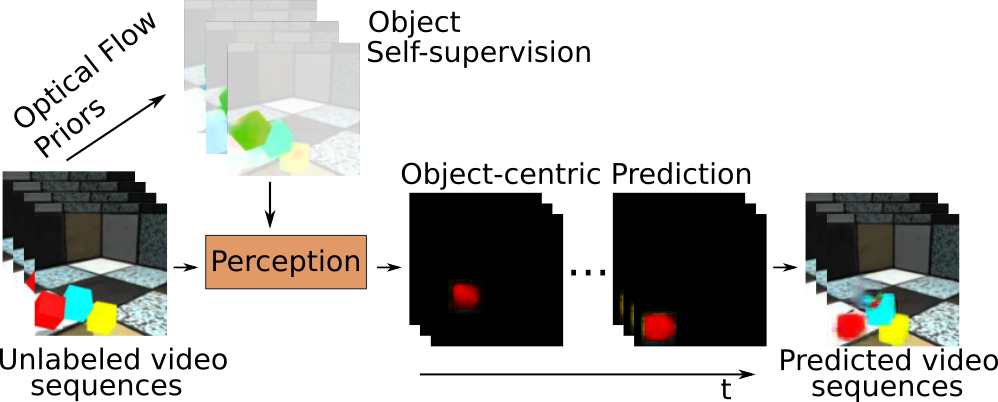}
\end{center}
   \caption{We seek to model physical interactions with a visual sensor and predict the future states of objects. During training our proposed approach leverages optical flow priors to generate object self-supervision. 
   We segment the image into objects, before predicting the future states of the objects, and using them to generate predictions of the next frames.
   In testing we predict future video sequences given a single frame as input.}
\label{fig:title}
\end{figure}

\begin{figure*}[h]
\begin{center}
\vspace{0.25cm}
\includegraphics[width=1\linewidth]{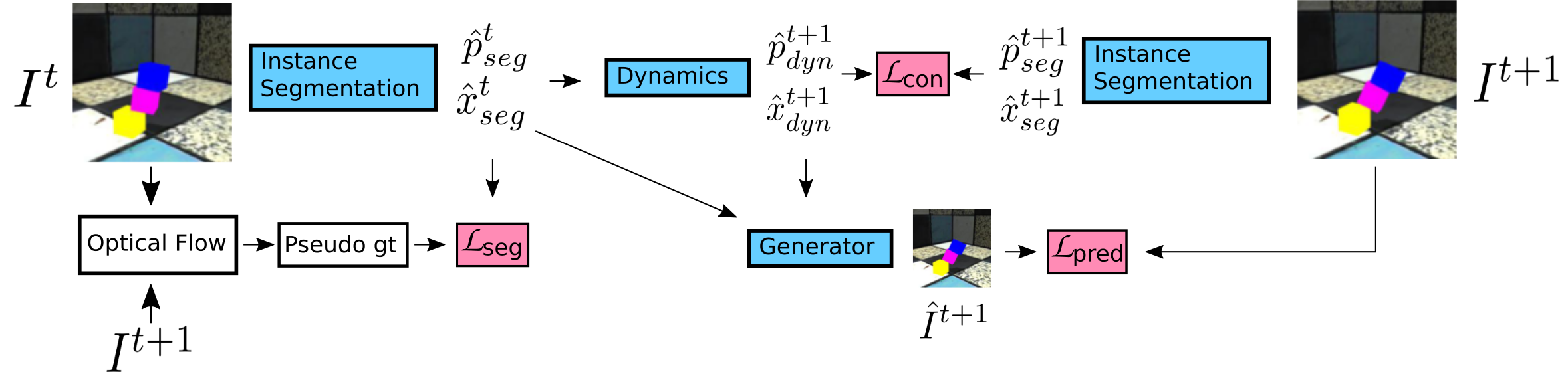}
\end{center}
   \caption{Overview of our object-centric predictive model for a single training time step. Given $\im^t$ we first extract patches $\segpatch^t$ and bounding boxes $\segbox^t$ from the instance segmentation module which is trained using our generated pseudo ground-truth. $\segpatch^t$ and $\segbox^t$ are then passed in the dynamics module that predicts future states $\dynpatch^{t+1}$ and $\dynbox^{t+1}$. We enforce consistency between the predicted future states and the outputs of the instance segmentation on frame $\im^{t+1}$. Finally, the future states are used as input to the generator that predicts the future image $\hat{\im}^{t+1}$. Blue color designates modules that are trained.}
\label{fig:pipeline}
\end{figure*}

Recent works have demonstrated that utilizing perceptual priors, via powerful computer vision models, reduces sample complexity, enables generalizability across environments, and largely increases performance in visuomotor tasks~\cite{sax2019learning,zhou2019does,mousavian2019visual,georgakis2019simultaneous}. 
Inspired by these methods, we make the observation that visual motion is a strong cue for objectness~\cite{xiong2018pixel} and propose a novel object-centric video predictive model that leverages state-of-the-art perception in the form of object instance segmentation and optical flow, and does not require object annotations. 
The perception model is trained end-to-end with dynamics and image generation models in order to predict a future frame sequence from a single input frame (see Figure~\ref{fig:title}).
The joint training allows for the perception model to fine-tune on the existing environment, while the dynamics and generation models benefit from the rich feature representation encoded in the perception model.
This results in an object-centric model that is not restricted only to environments where object level annotations are available, paving the way towards adaptable predictive models for manipulation tasks. Our contributions include: (a) the introduction of a novel prediction model that does not require object level annotations, (b) state-of-the-art results on the Shapestacks~\cite{groth2018shapestacks} dataset which demonstrate the benefits of our proposed method, and (c) the ability to fine-tune the perception model in new environments.

\section{Related Works}



\paragraph{Video Predictive Models} Learning to predict the future from raw observations such as images has gained large interest in recent years, especially in the context of learning representations for planning and robotic control~\cite{oh2015action,finn2017deep,chiappa2017recurrent,ebert2018visual,ebert2017self,finn2016unsupervised,schmeckpeper2020learning,levine2016end,zhang2019solar,rybkin2018learning}. 
In the seminal work of Finn et al.~\cite{finn2016unsupervised} an action-conditioned model that predicts future pixel motion was introduced. 
This was later demonstrated in~\cite{finn2017deep} to work with model predictive control for agents interacting with objects.
Ebert et al.~\cite{ebert2018visual} tackles the same problem but in a reinforcement learning setting, while~\cite{ebert2017self} attempts to learn a suitable representation to address occlusions in manipulation scenarios.
Other works improve the training of these models by leveraging both action-conditioned and action-free data, increasing the amount of data available~\cite{rybkin2018learning,schmeckpeper2020learning}.
Another line of works~\cite{lee2018stochastic,xue2016visual,denton2017unsupervised} are interested in predicting future frames not conditioned on a specific action, where the focus is on learning stochastic models to capture multiple possible futures of a dynamic scene.
For example, Lee et al.~\cite{lee2018stochastic} combined latent variational variable models with adversarially-trained models in order to produce both realistic and diverse future predictions. 
In contrast to all the aforementioned approaches that learn to predict future frames directly from pixels, we treat a scene as a collection of objects and learn how these objects appear in the future by modeling their physical interactions.


\paragraph{Object-centric Predictive Models} 
In order to address the high dimensionality of input pixels, several methods learn structured object-factorized representations for predicting either future images~\cite{ye2019compositional,janner2018reasoning,ye2019object,lerer2016learning,ehrhardt2020relate} or object properties~\cite{chang2016compositional,fragkiadaki2015learning,qi2020learning,watters2017visual,battaglia2016interaction}. The success of these methods correlates with their understanding of physical interactions and object dynamics in a scene. Byravan et al. learn an object centric prediction model from depth images~\cite{Byravan2017} and use it for robotic control~\cite{Byravan2018}. The work of Ye et al.~\cite{ye2019compositional} predicts future states of independent entities and learns to encode their interactions through a graph neural network. Similarly, Janner et al.~\cite{janner2018reasoning} follow an object factorization where object representations are organized as pairs before passed through a physics prediction network and a renderer. These object-centric models have been demonstrated in simple planning scenarios such as reaching~\cite{Byravan2018} and pushing objects~\cite{ye2019object}, but have also been illustrated in more complicated tasks such as stacking blocks~\cite{janner2018reasoning}, demonstrating that having access to an explicit object representation aids in planning. 
Several works in human video prediction rely on predicting the future state of a human skeleton before predicting the video, but these methods do not generalize to arbitrary objects~\cite{Cai2017,Fushishita2019}.

For predicting object properties, Fragkiadaki et al.~\cite{fragkiadaki2015learning} learns to predict object velocities in a simulated billiards game by individually modeling the temporal state of each ball. In Battaglia et al.~\cite{battaglia2016interaction}, a framework for learning about object relations called interaction networks is introduced, while the work of~\cite{watters2017visual} extends this framework to the visual domain.
More recently, Qi et al.~\cite{qi2020learning} explore the use of a region proposal network to extract object representations suitable for predicting future object locations and shapes.

Unlike our proposed method, all of these approaches require specific object related supervision such as object location, orientation, temporal association, velocities, or rely on strong assumptions such as the number of objects in the scene.


\section{Methodology}

We propose an object-centric video prediction pipeline that can operate from raw pixels.
Our pipeline is made up of three main components: an instance segmenter, which separates an object into entities, a dynamics model, which predicts the future location and state of each entity, and an image generator, which synthesizes the future frame from the predicted entities.
We describe each component in detail, as well as how the entire system is trained end-to-end.\footnote{All code is available at \url{https://github.com/kschmeckpeper/opa}} An overview of our pipeline is shown in Figure~\ref{fig:pipeline}.

\subsection{Instance Segmentation}
The first step in our object-centric video prediction method is to segment the input images into a set of entities.
We use the popular state-of-the-art instance segmentation method of Mask R-CNN~\cite{he2017mask} from the Detectron2~\cite{wu2019detectron2} implementation. 
Mask R-CNN is built on top of the Faster R-CNN~\cite{ren2015faster} object detection framework and predicts segmentation masks of detected object instances. The model uses a backbone (e.g., ResNet~\cite{he2016deep}) to extract image features maps that are initially passed to a region proposal network (RPN) which proposes candidate object bounding boxes. The network is able to compute object specific features through the RoIAlign layer which uses pooling operations to extract the image features corresponding to a specific region of interest. 
Then, the object features and bounding boxes are passed to separate branches of the network that are responsible to predict object labels, refine the candidate bounding boxes, and estimate the instance segmentation masks.
We chose this particular model because of its modularity and its ability to provide object specific features through region of interest pooling operations, but in principal any learnable instance segmentation method can be used in our pipeline. 
Since Mask R-CNN is typically trained on the COCO~\cite{lin2014microsoft} dataset, we devise a strategy to adapt it to novel robot manipulation environments. 

\begin{figure}[!t]
\begin{center}
\vspace{0.25cm}
\includegraphics[width=0.24\linewidth]{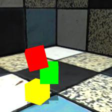}
\includegraphics[width=0.24\linewidth]{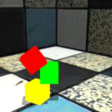}
\includegraphics[width=0.24\linewidth]{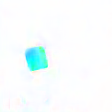}
\includegraphics[width=0.24\linewidth]{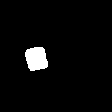}
\\
\vspace{0.5mm}
\includegraphics[width=0.24\linewidth]{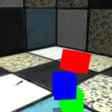}
\includegraphics[width=0.24\linewidth]{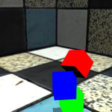}
\includegraphics[width=0.24\linewidth]{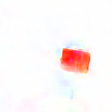}
\includegraphics[width=0.24\linewidth]{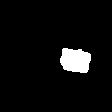}
\end{center}
   \caption{Pseudo ground-truth masks (last column) generated from optical flow (third column) of consecutive frames (first two columns). Even though this strategy only annotates moving objects, which are at the top of the stack, the instance segmentation learns a good representation and can generalize to static objects, as we show in our experiments. 
   }
\label{fig:pseudo_gt}
\end{figure}

As mentioned earlier, we assume that the environment we operate in does not offer any ground-truth at the object level. However, in video sequences motion cues often provide reliable location and mask information since all pixels belonging to a rigid object move in unison. To take advantage of this, we employ an optical flow method~\cite{Ranjan_CVPR_2017} on consecutive frames of our videos. 
The magnitude of the predicted flow is thresholded to keep any image areas that have moved more than $1\%$ of the image dimensions. The remaining pixels are organized into connected components in image space. The convex hull of each connected component is then taken as a pseudo ground-truth mask and all masks are defined to belong to a single object class. Figure~\ref{fig:pseudo_gt} shows examples of generated pseudo ground-truth masks from optical flow. The idea is to learn what \textit{can} move in a scene as that is an inherent indication of objectness~\cite{xiong2018pixel,dave2019towards}. This also allows us to treat everything else in the scene, such as the floor tiles, as background.  


During training we use the generated ground-truth to estimate the Mask R-CNN losses $L_{cls}$, $L_{box}$, $L_{mask}$ 
and the RPN losses $L_{obj}$, $L_{reg}$. 
$L_{cls}$ is a classification log loss, $L_{box}$ is a smooth $L_1$ loss for bounding box regression, and $L_{mask}$ is a per-pixel binary cross-entropy loss. 
Regarding RPN, $L_{obj}$ is a binary log loss, and $L_{reg}$ is identical to $L_{box}$.
More details can be found in~\cite{he2017mask,ren2015faster}.
We define $L_{seg}$ in our model to be the summation of all these losses.
In practice, we pre-train the instance segmentation model on a small set of training sequences before integrating with the rest of the components.


In the forward pass of the model, given an image $\im$ the instance segmentation extracts bounding boxes, $\segbox$, and patches, $\segpatch = (\segfeat, \segmask)$ for each detected object instance, where $\segfeat$ is the feature representation of the object (from RoIAlign) and $\segmask$ is the mask defined as a foreground probability of each pixel location.





\subsection{Dynamics Prediction}
Given a set of bounding boxes, $\segbox$, and patches, $\segpatch$, the dynamics model, $\dynmodel$, seeks to predict their state at the next timestep:

\begin{equation}
    \dynbox^{t+1}, \dynpatch^{t+1} = \dynmodel(\segbox^{t}, \segpatch^{t})
\end{equation}

Each bounding box and patch is encoded to form a latent representation.
The latent representation is passed through a fully-connected neural network to generate the latent encoding at the next time step.
The predicted encodings are decoded to get the predicted future location of the bounding box, $\dynbox^{t+1}$ and its patch, $\dynpatch^{t+1}$.

During training, the algorithm has access to the ground truth images from future timesteps, so we can impose a consistency loss between the entities predicted by the dynamics model and the entities found by running the instance segmentation model on the future frame:

\begin{equation}
\begin{split}
    \mathcal{L}_{con} = & \left\| \dynmask^{t+1} \odot \dynfeat^{t+1} - \segmask^{t+1} \odot \segfeat^{t+1} \right\|^2 + \\
    & \left\| \dynbox^{t+1} - \segbox^{t+1} \right\|^2
    \label{eqn:consistency_loss}
\end{split}    
\end{equation}

To ensure that the consistency loss is applied to corresponding entities, we associate the instance predictions from the input image $\im^t$ to those of future image $\im^{t+1}$ based on bounding box centroid proximity in image space.
This procedure is repeated for any following future images and establishes a constant number of entities throughout a single sequence.
Another option for establishing these associations is to warp $\segbox$ and $\segpatch$ according to optical flow between $\im^t$ and $\im^{t+1}$. However, we discovered that this was unreliable for long sequences due to artifacts introduced during warping.

\subsection{Image Generation}
The final image is generated as a function of the previous image, $\im^{t}$, the previous bounding boxes $\segbox^{t}$ and segmentation masks $\segmask^{t}$, and the predicted future patches $\dynpatch^{t+1}$ and bounding boxes $\dynbox^{t+1}$:
\begin{equation}
    \predim^{t+1} = \generator(\im^{t}, \segbox^{t}, \segmask^{t}, \dynbox^{t+1}, \dynpatch^{t+1}).
\end{equation}

The image generator $\generator$ works by first generating separate images containing pixels for the predicted objects $\hat{I}_{obj}$, pixels for the background $\hat{I}_{back}$, synthetic pixels $\hat{I}_{synth}$, and then compositing them together. We will describe how each image is created and how they are combined in the subsequent paragraphs.
First, we define a transformation function $\gamma$, which converts the segmentation masks $\hat{m}^t$ to image coordinates using their corresponding bounding boxes $\hat{x}^t$:

\begin{equation}
    \hat{M}^t = \gamma(\hat{m}^{t}, \hat{x}^{t})
\end{equation}
where $\hat{M}^t$ is a binary mask with the same dimensions as the previous image $I^t$.


The contribution of the objects is made by passing each patch through a convolutional neural network $\feattopix$ that decodes from features to pixels, and then taking the sum of the pixels multiplied by their corresponding masks:
\begin{equation}
    \objim^{t+1} = \sum^n_{i=1} \gamma \left(\dynmaski^{t+1} \odot \feattopix(\dynpatchi^{t+1}), \dynbox^{t+1}\right)
\end{equation}
where $n$ is the number of object instances.

Our object-centric video prediction model assumes that any change in the image can be explained by an object, meaning the background pixels, which are copied directly between frames, are all pixels that are not objects.
The background mask is therefore found by subtracting all current and previous object masks from a mask of all ones:
\begin{equation}
    \hat{M}_{back}^{t+1} = 1 - \sum^n_{i=1} \hat{M}_{seg,i}^{t} - \sum^n_{i=1} \hat{M}_{dyn,i}^{t+1}.
\end{equation}
The background pixels are then the product of the background mask and the image from the previous timestep:
\begin{equation}
    \backim^{t+1} = \hat{M}_{back}^{t+1} \odot \im^{t}.
\end{equation}

We generate synthetic pixels to fill in the holes between the background mask and the object masks.  These are primarily regions of the image that were occluded by an object but are no longer occluded.  The mask that controls the locations of the synthetic pixel is given by the following equation:
\begin{equation}
    \hat{M}_{synth}^{t+1} = 1 - \hat{M}_{back}^{t+1} - \sum^n_{i=1} \hat{M}_{dyn,i}^{t+1}.
\end{equation}

The synthetic pixels are generated from a convolutional neural network $\Psi$ that takes the object pixels, $\objim$, and the background pixels, $\backim$, as input:
\begin{equation}
    \synthim^{t+1} = \hat{M}_{synth}^{t+1} \odot \Psi(\objim^{t+1}, \backim^{t+1}).
\end{equation}

\begin{figure}
    \centering
    \vspace{0.25cm}
    \includegraphics[width=0.8\columnwidth]{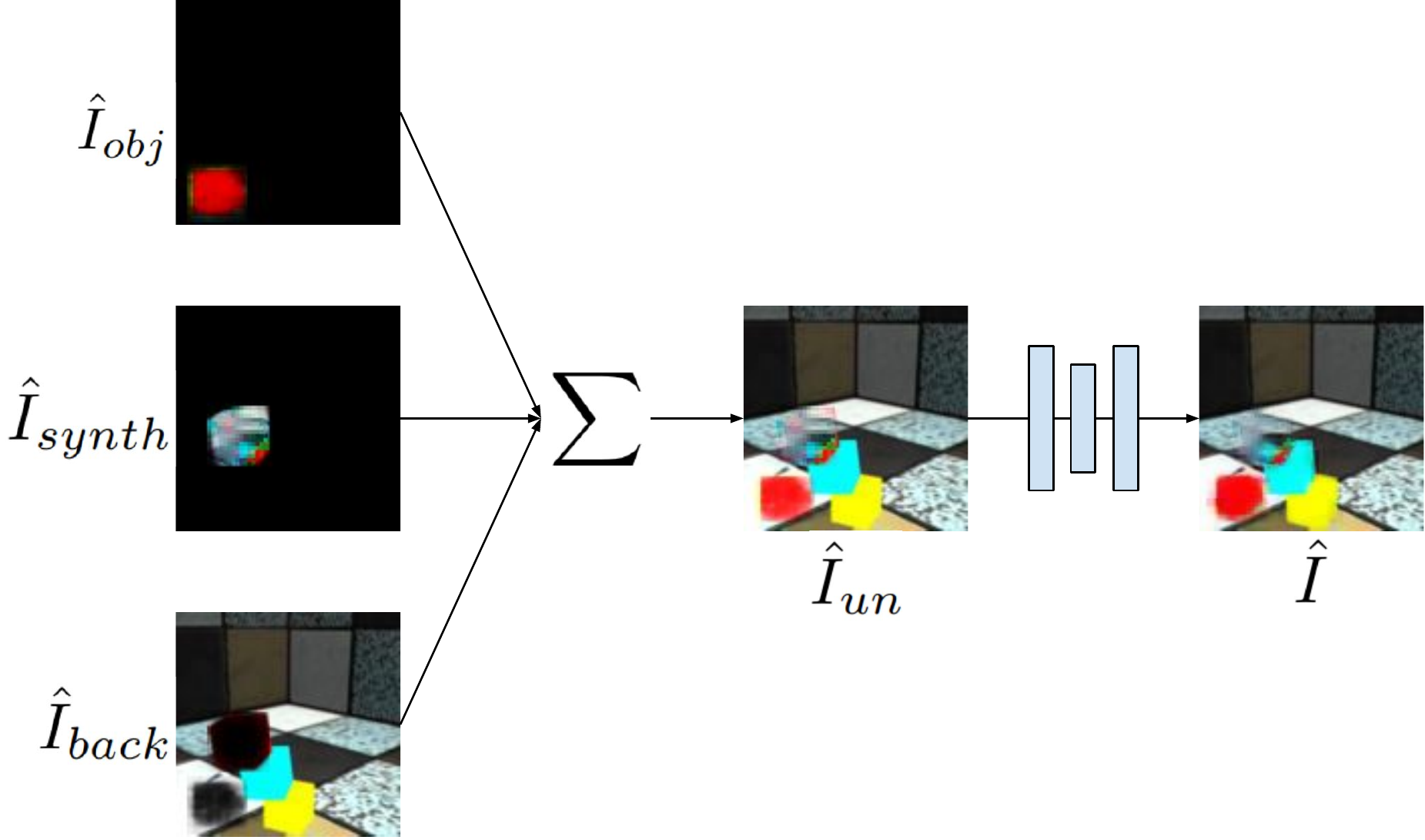}
    \caption{Image generation.  This shows the components of the image generator $\generator$. The unrefined image, $\unim$, is generated by summing together the contributions from the objects, $\objim$, the background, $\backim$, and the synthetic pixels, $\synthim$.  The unrefined image is passed through a neural network to clean up the edges between its component patches, generating the final image, $\predim$.}
    \label{fig:image_gen}
\end{figure}
We then sum together the background pixels, $\backim^{t+1}$, the object pixels, $\objim^{t+1}$, and the synthetic pixels, $\synthim^{t+1}$, generating an initial unrefined image, $\unim^{t+1}$:

\begin{equation}
    \unim^{t+1} = \backim^{t+1} + \objim^{t+1} + \synthim^{t+1}.
\end{equation}

The unrefined image, $\unim$, is passed through a convolutional neural network to clean up the edges between its component patches, generating the final image, $\predim$.  The image generator process is shown in Figure~\ref{fig:image_gen}.
The final predicted image and the unrefined image are supervised using an L2 loss with the ground truth future image:   
\begin{equation}
\begin{split}
    \mathcal{L}_{pred} = &\left\| \predim^{t+1}- \im^{t+1} \right\|^2 + \left\| \unim^{t+1}- \im^{t+1} \right\|^2\\
    &+\alpha \left\| \binarymask \odot \predim^{t+1}- \binarymask \odot \im^{t+1} \right\|^2.
    \label{eqn:prediction_loss}
\end{split}
\end{equation}
The third term is a masking loss that provides attention to foreground pixels,
where $\binarymask$ are the predicted binary masks, and $\alpha$ is a weighting hyperparameter.
Both the final predicted image and the unrefined predicted image are supervised to ensure that the revising network does not attempt to learn dynamics.

The entire system is jointly optimized to minimize the combined loss:
\begin{equation}
    \mathcal{L} = \mathcal{L}_{pred} + c_1 \mathcal{L}_{con} + c_2 \mathcal{L}_{seg}
\end{equation}
where $c_1$, and $c_2$ are hyperparamters that weight the relative importance of each part of the loss.


\section{Experiments}

\newcommand{\predwidth}{0.10\textwidth}

\begin{figure*}
    \centering
    \vspace{0.25cm}
    \rotatebox{90}{\scriptsize{GT}}
    \includegraphics[width=\predwidth]{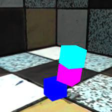}
    \includegraphics[width=\predwidth]{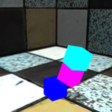}
    \includegraphics[width=\predwidth]{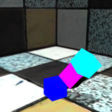}
    \includegraphics[width=\predwidth]{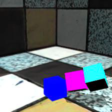}
    \rotatebox{90}{\scriptsize{GT}}
    \includegraphics[width=\predwidth]{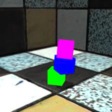}
    \includegraphics[width=\predwidth]{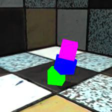}
    \includegraphics[width=\predwidth]{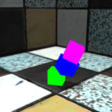}
    \includegraphics[width=\predwidth]{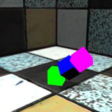}
      
    \rotatebox{90}{\scriptsize{CVP\cite{ye2019compositional}}}
    \includegraphics[width=\predwidth]{env_blocks-easy-h=3-vcom=1-vpsf=0-v=6_context_frames_00.png}
    \includegraphics[width=\predwidth]{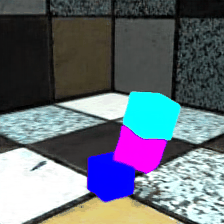}
    \includegraphics[width=\predwidth]{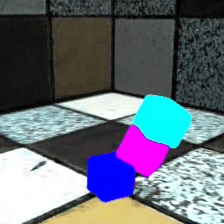}
    \includegraphics[width=\predwidth]{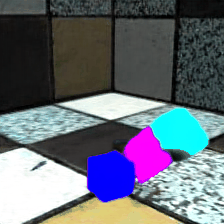}
    \rotatebox{90}{\scriptsize{CVP\cite{ye2019compositional}}}
    \includegraphics[width=\predwidth]{env_blocks-easy-h=3-vcom=1-vpsf=0-v=156_context_frames_00.png}
    \includegraphics[width=\predwidth]{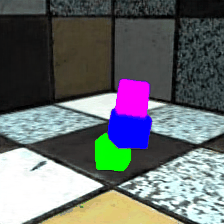}
    \includegraphics[width=\predwidth]{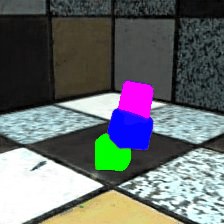}
    \includegraphics[width=\predwidth]{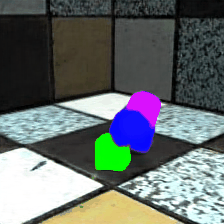}

    \rotatebox{90}{\scriptsize{No-Factor\cite{lerer2016learning}}}
    \includegraphics[width=\predwidth]{env_blocks-easy-h=3-vcom=1-vpsf=0-v=6_context_frames_00.png}
    \includegraphics[width=\predwidth]{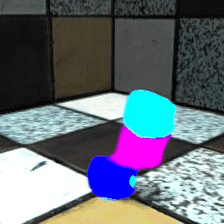}
    \includegraphics[width=\predwidth]{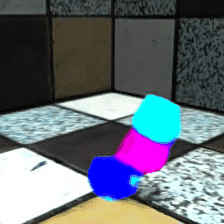}
    \includegraphics[width=\predwidth]{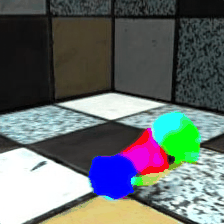}
    \rotatebox{90}{\scriptsize{No-Factor\cite{lerer2016learning}}}
    \includegraphics[width=\predwidth]{env_blocks-easy-h=3-vcom=1-vpsf=0-v=156_context_frames_00.png}
    \includegraphics[width=\predwidth]{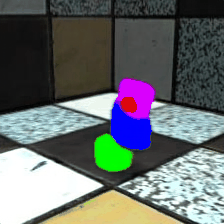}
    \includegraphics[width=\predwidth]{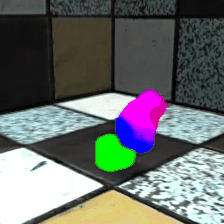}
    \includegraphics[width=\predwidth]{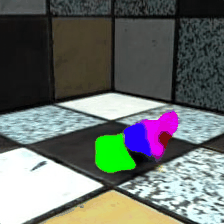}

    \rotatebox{90}{\scriptsize{OPA (Ours)}}
    \includegraphics[width=\predwidth]{env_blocks-easy-h=3-vcom=1-vpsf=0-v=6_context_frames_00.png}
    \includegraphics[width=\predwidth]{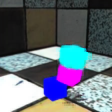}
    \includegraphics[width=\predwidth]{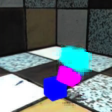}
    \includegraphics[width=\predwidth]{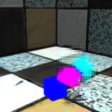}
    \rotatebox{90}{\scriptsize{OPA (Ours)}}
    \includegraphics[width=\predwidth]{env_blocks-easy-h=3-vcom=1-vpsf=0-v=156_context_frames_00.png}
    \includegraphics[width=\predwidth]{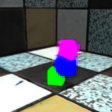}
    \includegraphics[width=\predwidth]{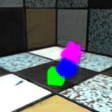}
    \includegraphics[width=\predwidth]{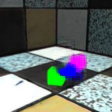}
    
    \vspace{5mm}
    
    \rotatebox{90}{\scriptsize{GT}}
    \includegraphics[width=\predwidth]{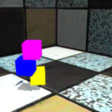}
    \includegraphics[width=\predwidth]{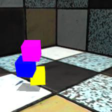}
    \includegraphics[width=\predwidth]{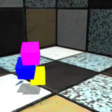}
    \includegraphics[width=\predwidth]{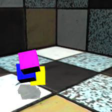}
    \rotatebox{90}{\scriptsize{GT}}
    \includegraphics[width=\predwidth]{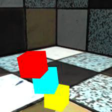}
    \includegraphics[width=\predwidth]{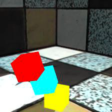}
    \includegraphics[width=\predwidth]{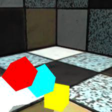}
    \includegraphics[width=\predwidth]{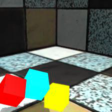}

    \rotatebox{90}{\scriptsize{CVP\cite{ye2019compositional}}}
    \includegraphics[width=\predwidth]{env_blocks-hard-h=3-vcom=1-vpsf=0-v=66_context_frames_00.png}
    \includegraphics[width=\predwidth]{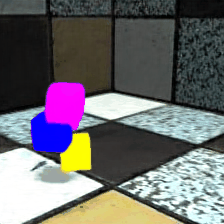}
    \includegraphics[width=\predwidth]{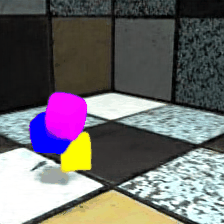}
    \includegraphics[width=\predwidth]{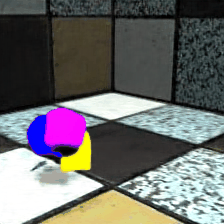}
    \rotatebox{90}{\scriptsize{CVP\cite{ye2019compositional}}}
    \includegraphics[width=\predwidth]{env_blocks-hard-h=3-vcom=1-vpsf=0-v=127_context_frames_00.png}
    \includegraphics[width=\predwidth]{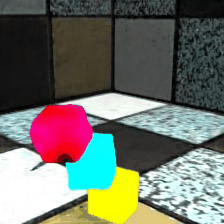}
    \includegraphics[width=\predwidth]{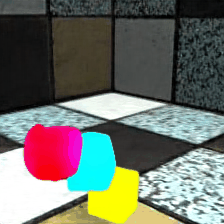}
    \includegraphics[width=\predwidth]{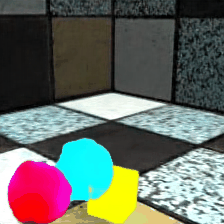}
    
    \rotatebox{90}{\scriptsize{No-Factor\cite{lerer2016learning}}}
    \includegraphics[width=\predwidth]{env_blocks-hard-h=3-vcom=1-vpsf=0-v=66_context_frames_00.png}
    \includegraphics[width=\predwidth]{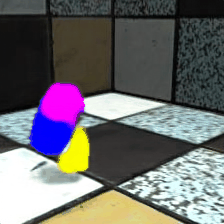}
    \includegraphics[width=\predwidth]{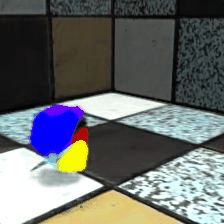}
    \includegraphics[width=\predwidth]{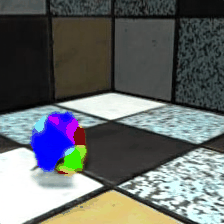}
    \rotatebox{90}{\scriptsize{No-Factor\cite{lerer2016learning}}}
    \includegraphics[width=\predwidth]{env_blocks-hard-h=3-vcom=1-vpsf=0-v=127_context_frames_00.png}
    \includegraphics[width=\predwidth]{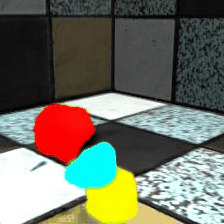}
    \includegraphics[width=\predwidth]{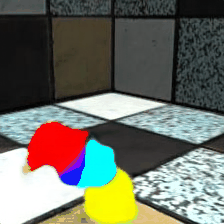}
    \includegraphics[width=\predwidth]{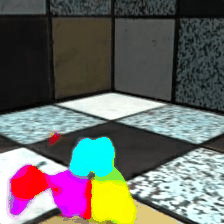}

    \rotatebox{90}{\scriptsize{OPA (Ours)}}
    \includegraphics[width=\predwidth]{env_blocks-hard-h=3-vcom=1-vpsf=0-v=66_context_frames_00.png}
    \includegraphics[width=\predwidth]{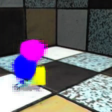}
    \includegraphics[width=\predwidth]{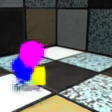}
    \includegraphics[width=\predwidth]{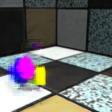}
    \rotatebox{90}{\scriptsize{OPA (Ours)}}
    \includegraphics[width=\predwidth]{env_blocks-hard-h=3-vcom=1-vpsf=0-v=127_context_frames_00.png}
    \includegraphics[width=\predwidth]{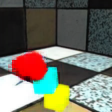}
    \includegraphics[width=\predwidth]{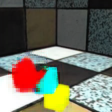}
    \includegraphics[width=\predwidth]{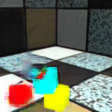}
    
    \newcommand{\capwidth}{1.5cm}
    Input\hspace{\capwidth}   t=1\hspace{\capwidth}  t=2\hspace{\capwidth} t=3\hspace{\capwidth}  Input\hspace{\capwidth} t=1\hspace{\capwidth} t=2\hspace{\capwidth} t=3
    \caption{Qualitative Prediction Results.  OPA, our prediction model, is able to achieve better performance than the non-object-centric prediction model from \cite{lerer2016learning}.  Our predicted videos retain more consistent object properties, such as color and shape than the competing approach.  Additionally, OPA is able to achieve comparable results to CVP\cite{ye2019compositional}, despite CVP requiring ground truth annotations including bounding boxes and object tracks.  OPA is able to bring the predictive quality of object-centric video prediction to datasets without annotations.}
    \label{fig:qualitative_results}
    \vspace{-0.5cm}
\end{figure*}

We seek to show that our prediction model can provide good video prediction when trained without any annotations, such as bounding boxes, on the training data.
In order to do this, the model must be able to segment the input images and predict the future states of objects. 
We present video prediction results, and evaluate the internal object detection and object-centric prediction.

\noindent \textbf{Implementation Details:} We use the Mask R-CNN model with a ResNet50+FPN backbone that is pretrained on the COCO~\cite{lin2014microsoft} dataset. Our dynamics model $\dynmodel$ is implemented as a 4-layer fully connected network, the feature-to-pixel network $\feattopix$ is a 3-layer convolutional network and the synthetic pixel generator network, $\Psi$ is a stacked-hourglass network.  Our patch size is 14 pixels. For more information, see the official implementation.


\begin{table*}[h]
    \centering
    \vspace{0.25cm}
    \begin{tabular}{|c|c|c|}
\hline
         Method & MSE ($\downarrow$) & LPIPS \cite{zhang2018perceptual} ($\downarrow$)  \\
\hline
         CVP \cite{ye2019compositional} (Requires GT bounding boxes) & $0.010134 \pm 0.000318 $ & $ 0.051941 \pm 0.001432$ \\
         No-factor\cite{lerer2016learning} (w/ GT bounding boxes) & $0.012974 \pm 0.000327$ & $0.067548 \pm 0.001381$ \\
\hline
\hline
        No-factor\cite{lerer2016learning} (w/o GT bounding boxes) & $0.010821 \pm 0.000341$ & $0.069826 \pm 0.001524$ \\
        OPA (Ours) (w/o GT bounding boxes) & $\mathbf{0.009720 \pm 0.000314}$ & $\mathbf{0.0619602 \pm 0.001567}$ \\
\hline
    \end{tabular}
    \caption{Mean and standard error for prediction on shapestacks.}
    \label{tab:prediction_error}
    \vspace{-0.75cm}
\end{table*}

\subsection{Video Prediction}
We perform our video prediction experiments in an extension of the Shapestacks dataset \cite{groth2018shapestacks,ye2019compositional}.  The environment is made up of an unstable stack of blocks placed in front of the camera.  In all experiments, the model is given a single starting frame and must predict the subsequent images.  We sample every other frame of the original dataset in order to speed up the videos.

We compare our approach to several existing approaches.
Compositional Video Prediction (CVP), a state of the art object-centric video prediction that requires ground truth bounding boxes at each timestep \cite{ye2019compositional}.
We additionally compare against the non-object centric approach, which we train both with and without access to the ground-truth bounding boxes\cite{lerer2016learning}.
In comparison, our approach requires no annotations or ground-truth bounding boxes, allowing it to learn directly from pixels.
For all baselines, we use the implementations provided by~\cite{ye2019compositional}.

The results of the video prediction experiments are shown in Table~\ref{tab:prediction_error}. We evaluate the prediction performance across sequences of length three, given a single input frame, using the Mean Square Error (MSE) and the Learned Perceptual Image Patch Similarity (LPIPS)\cite{zhang2018perceptual} metrics.  Qualitative results are shown in Figure~\ref{fig:qualitative_results}.

Our model is able to achieve better prediction results than the non-object-centric no-factor prediction model~\cite{lerer2016learning}.  
Unlike~\cite{lerer2016learning} which tackles this problem by learning to map pixels to pixels in an object-agnostic manner,
our approach is able to maintain more consistent object properties, such as color and shape. 

We treat CVP's performance as the lower bound error that an object-centric model can achieve because it makes use of all possible object annotations during training. We note that our performance is comparable even though 
we train without ground truth bounding box annotations or object tracks, making it so our approach is much more widely applicable.

\subsection{Learned Segmentation}

\begin{figure}[!t]
\begin{center}
\includegraphics[width=0.24\linewidth]{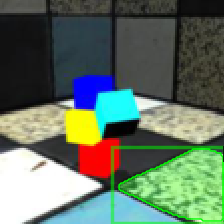}
\includegraphics[width=0.24\linewidth]{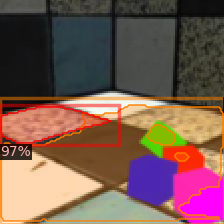}
\includegraphics[width=0.24\linewidth]{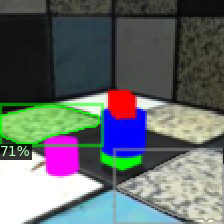}
\includegraphics[width=0.24\linewidth]{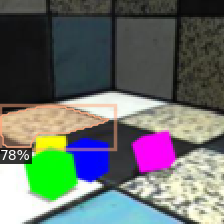}
\\
\vspace{0.5mm}
\includegraphics[width=0.24\linewidth]{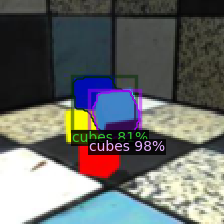}
\includegraphics[width=0.24\linewidth]{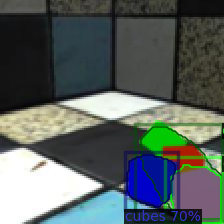}
\includegraphics[width=0.24\linewidth]{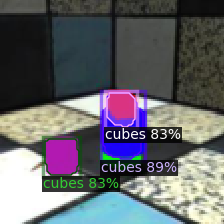}
\includegraphics[width=0.24\linewidth]{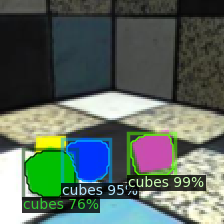}
\end{center}
   \caption{Mask R-CNN detections before (top row) and after (bottom row) the instance segmentation model was finetuned using our generated pseudo ground-truth.}
\label{fig:maskrcnn_detections}
\end{figure}

We demonstrate that our proposed approach of generating pseudo ground-truth from optical flow cues provides sufficient supervision to fine-tune the instance segmentation model. Figure~\ref{fig:maskrcnn_detections} illustrates Mask R-CNN detections before and after our pseudo ground-truth training strategy. 
Notice how the model generalizes to multiple objects in each scene even though the pseudo ground-truth from optical flow usually only annotates objects at the top of the stack. 

The fact that optical flow priors~\cite{Ranjan_CVPR_2017} can provide supervision to the instance segmentation model is actually a confirmation
that visual motion can provide strong objectness cues~\cite{xiong2018pixel,dave2019towards}, even though it was pre-trained using a significantly different dataset. 
In contrast, Mask R-CNN, a powerful object instance segmentation model, requires finetuning in the context of our experimental setup.


\subsection{Object-Centric prediction}

\newcommand{\ocwidth}{0.2\columnwidth}
\begin{figure}
    \centering
    
    
    
    \rotatebox{90}{\scriptsize{\hspace{0.2cm}  Ground Truth}}
    \includegraphics[width=\ocwidth]{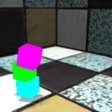}
    \includegraphics[width=\ocwidth]{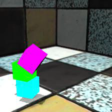}
    \includegraphics[width=\ocwidth]{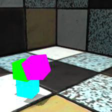}
    \includegraphics[width=\ocwidth]{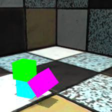}
    
    \rotatebox{90}{\scriptsize{Pred. Object Patch}}
    \includegraphics[width=\ocwidth]{env_blocks-easy-h=3-vcom=2-vpsf=0-v=16_context_frames_00.png}
    \includegraphics[width=\ocwidth]{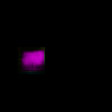}
    \includegraphics[width=\ocwidth]{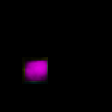}
    \includegraphics[width=\ocwidth]{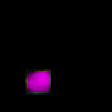}
    
    \rotatebox{90}{\scriptsize{Predicted Image}}
    \includegraphics[width=\ocwidth]{env_blocks-easy-h=3-vcom=2-vpsf=0-v=16_context_frames_00.png}
    \includegraphics[width=\ocwidth]{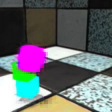}
    \includegraphics[width=\ocwidth]{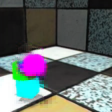}
    \includegraphics[width=\ocwidth]{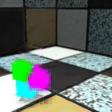}
    
    \newcommand{\capwidtho}{1.5cm}
    Input\hspace{\capwidtho}   t=1\hspace{\capwidtho}  t=2\hspace{\capwidtho} t=3
    
    \caption{Object-centric prediction.  Our model predicts the future states of individual objects, middle.  By predicting the states of multiple objects and reasoning about non-object pixels, our system is able to predict the full image, bottom.}
    \label{fig:object_centric_objects}
\end{figure}

Additionally, our prediction pipeline maintains good object representations during prediction.  An example of the predicted object is shown in Figure~\ref{fig:object_centric_objects}.
Being able to predict the future states of individual objects is important, not just because it improves the quality of the final predicted image, but also because it allows for easier interfacing with downstream tasks.
Many robotic tasks involve interacting with objects, and maintaining an explicit object representation allows the agent to plan on those representations, rather than having to plan in pixel space.







\section{Conclusions}

We present Object-centric Prediction without Annotation (OPA), an approach to training object-centric video prediction models purely from unlabeled video. We have demonstrated our model's advantage towards object-agnostic prediction models and have shown comparable performance to methods that use dense object annotations during training. 
Object-centric video prediction models offer a promising way to allow robots to learn about the dynamics of the world from cheap video data.
They have the ability to handle more difficult dynamics while providing better interfaces for robotic control than pixel-to-pixel prediction models.  OPA's ability to learn object-centric video prediction from video without annotations offers the promise of bringing object-centric video prediction to more applications, allowing robots to better understand and anticipate their environments.

\section*{Acknowledgements}
Research was sponsored by the Army Research Office and was accomplished under grants ARO MURI W911NF-20-1-0080, NSF CPS 2038873, ARL DCIST CRA W911NF-17-2-0181, ONR N00014-17-1-2093, and by the Honda Research Institute.

\bibliographystyle{./IEEEtran} 
\bibliography{./IEEEexample}

\end{document}